\documentclass[11pt]{article}
\usepackage{times}
\usepackage{url}
\usepackage{latexsym}
\usepackage{rotating}
\usepackage{graphicx}
\usepackage{natbib}
\newlength\titlebox 
\title{Detection is the central problem in real-word spelling correction}
\date{August 13, 2014}
\author{
  L. Amber Wilcox-O'Hearn \\
  Department of Computer Science \\
  University of Toronto \\ 
  {\tt amber@cs.toronto.edu} \\
  }

\begin{document}
\maketitle
\begin{abstract}
Real-word spelling correction differs from non-word spelling correction in its aims and its challenges.
Here we show that the central problem in real-word spelling correction is \emph{detection}.
Methods from non-word spelling correction, which focus instead on selection among candidate corrections, do not address detection adequately,
because detection is either assumed in advance or heavily constrained.

As we demonstrate in this paper, merely discriminating between the intended word and a random close variation of it within the context of a sentence is a task that can be performed with high accuracy using straightforward models.
Trigram models are sufficient in almost all cases.
The difficulty comes when every word in the sentence is a potential error, with a large set of possible candidate corrections.
%By allowing a more realistically open set of possible errors than in most previous research, we show that 

Despite their strengths, trigram models cannot reliably find true errors without introducing many more, at least not when used in the obvious sequential way without added structure.
The detection task exposes weakness not visible in the selection task.
%This is true even if probable words are excluded from consideration in advance.
\end{abstract}

\section{Introduction: real-word spelling correction}

%\subsection{Non-word vs. real-word spelling correction}
The task we typically think of as spelling correction corresponds to the action we take when we are reading and encounter a word that we do not recognise.
Such a word presents an immediate, unambiguous problem.
It is possible that this is a new word that we do not know, and that is why we do not recognise it.
However, it could instead be a word we do know that has been misspelled, or spelled in a way we have not seen before.
If we decide it is indeed a misspelling,
it is usually because we have thought of another word that could go in its place
%fits in the context and 
close enough in spelling to what was written that it is a plausible mistake.
%Is it a new word, or is it a misspelling or spelling variation of a word we do know?
% While reading or listening to speech, we have expectations about what words are possible at a given point in a sentence.
%If we decide the latter, we then need to propose a replacement from a list of known words.
Trying to find a replacement for an unrecognised word is the problem of \emph{non-word spelling correction}.

In natural language processing software, except for in applications specifically designed for editing, non-word spelling correction is usually not addressed at all\footnote{
Information retrieval is a notable exception, as search engines usually attempt to correct both real-word and non-word spelling errors.
}.
Rather than assuming an unknown word is an error, it is assumed to be a valid new word.
That is because natural vocabularies are always unbounded, and texts normally have many single occurrences of rare and new words, especially names.
NLP systems therefore expect to encounter words they didn't have available in training.
Of course, some of these words may turn out to be spelling errors or variations of words that the system already knew about
\footnote{In future work, we may incorporate the decision of whether to treat such words as new, or instead correct or normalise them.
In this work, however, we do not attempt non-word spelling correction, and always treat unknown words as correct.}.

The focus of this paper is what happens when 
we encounter a known word that is sufficiently unlikely in its context that it is worth wondering whether the intended word was actually not the one we have observed.
%we may encounter a known word that is sufficiently unlikely in its context that it is worth wondering whether the intended word was actually not the one we have observed.
This problem is called \emph{real-word spelling correction}, or, sometimes, \emph{context-sensitive spelling correction}.
Real-word spelling correction requires different methods from non-word spelling correction.

\subsection{Detection vs. correction}

Detection, that is, realising that there has been an error, is trivial in the non-word problem.
We detect a non-word error only if the word we encounter is not in our vocabulary.
In this case, we may still decide to classify the word as a new word, and not an error.
In spelling correction applications, which typically run interactively, this decision is delegated to the user.
As just discussed, other applications normally simply assume that an unseen word is a new word type.

For the real-word case, however, detection necessarily involves having some model of what we expect the text to be like, so that we can tell whether those expectations have been violated.
We assume that we have detected a real-word spelling error when the probability of the observed word is sufficiently low that it undermines our confidence that the writer intended it.

This suggests an anomaly-detection approach, in which correction is attempted only if the probability of a text falls outside of the expected variation in probability.
As far as we know, that approach has not been tried 
\footnote{Our experiments with this approach forthcoming}.
Instead, detection of real-word errors is a by-product of correction.
Typically, there is a subset of the vocabulary of words considered a priori to be possible candidates for correction.
Whenever one of those words is encountered, we consider correcting it, and retain the correction if it makes the text sufficiently more probable than the original.
This means that in the real-word case as typically approached, detection itself is dependent on decisions that are ultimately part of the \emph{correction} process.

Correction itself has two steps.
First, a set of alternatives must be generated.
Then one (or more) of the alternatives must be selected.
In the non-word task, alternatives are generated by searching the dictionary for known words that are close in edit distance.
Again, because non-word spelling correction has traditionally been employed only in interactive applications specifically for aiding a writer with spelling, selection has been left to the user.

\section{Previous approaches}

\subsection{Generation}
For real-word spelling correction,
the generation task is usually construed as creating \emph{confusion sets}
--- sets of words that are often mistakenly interchanged,
such as \{`to', `two', `too'\}.
When a word is encountered that belongs to one of the known confusion sets, all of the other members of the set are considered candidate corrections.
Most previous methods have relied on pre-defined and typically hand-crafted confusion sets.

Using pre-defined confusion sets implicitly makes \emph{selection} the task of interest, casting it as a classification problem.
That has the advantage of allowing specific features to be learned for discrimination, but it also has disadvantages.
It is resource intensive:
It requires either human expertise about what errors are possible or a repository of errors and their corrections to be learned from.
Moreover, it suffers from lack of generality, both in the errors it can detect and correct, and in being language specific.

A more flexible approach to generation is to algorithmically define confusion sets based on some measureable characteristic.
The use of edit distance in non-word spelling correction is an example of this.
Any property of words that allows a distance comparison to be made between words could serve.
\citet{toutanova2002pronunciation} use a phonetic distance for non-word spelling correction.
\citet{mays1991context} use Damerau-Levenshtein edit distance (hereafter \emph{DL-distance}) to generate candidate corrections for real-word spelling errors.
The work presented here is based on the methods in that paper.
In particular, we generate confusion sets in the same way.

A major advantage of constructing confusion sets algorithmically is that in principle errors can be detected in any word.
This flexibility, however, comes with more opportunity to ``detect'' errors where they haven't actually occurred.

\subsection{Selection}
The task of selection starts with deciding which of the alternatives best fits in the context.
With pre-defined confusion sets, supervised machine learning techniques apply.
For example, 
\citet{golding1999winnow} use a Winnow-based algorithm to learn to discriminate among the words in their confusion sets according to surrounding words.

With algorithmically defined confusion sets, the space of distinctions to learn is bounded only by the vocabulary.
Contextual distinctions for essentially every word in the vocabulary are needed;
at least distinctions among all words that can ever fall within the same confusion set.
Note that unlike a typical pre-defined confusion set, algorithmic confusion sets may overlap and are not necessarily symmetric or transitive.
For example, if we take a Levenshtein edit distance of $1$ as a criterion for confusability,
`as' will be in the confusion set derived from both `a' and `ask', but `a' is not in the confusion set of `ask'.

One commonly used statistical model that represents every word in the vocabulary is the n-gram model.
We follow 
\citet{mays1991context} in using trigrams for selection.
Another kind of word-centric statistical model of contextual fit that could be used is the topic model.
%It could be used by itself or, because the contextual representation is independent of that used for n-grams,
%it seems likely that it could be used profitably in combination with them.
We leave this to future work.

As mentioned above, in our model of real-word spelling correction, the detection decision is folded into the correction process.
This adds two more selection constraints in addition to contextual fit.
Because correction presumes that an error has been made, we want to be biased toward selecting the original word, that is, not correcting at all.
Moreover, we may want to distinguish among the possible selections based on how plausible it is that each one was intended as the original.
Again following \citet{mays1991context}, we use a noisy channel model to represent these multiple constraints.

\subsection{The noisy channel model}
To frame spelling correction as a noisy channel problem, we treat the observed sentence as a signal, $S$, which has passed through a noisy channel (the typist).
This channel might have introduced errors into the sentence.
Our task is to find the most likely original signal $S'$ (the intended sentence, generated by a language model).
The probability that the typist types a word correctly is a parameter ${\alpha}$, which is the same for all words.

For each word, the remaining probability mass ($1 - {\alpha}$), the probability that the word is mistyped as another real word, is distributed equally among all its spelling variations.
This implicitly values as zero the probability that a word was transformed by the channel into something other than a variation considered by our model.
We reconcile this by letting the probability assigned to words in our confusion sets include the possibility that the original word was transformed into a non-word, or a more distant variation, and then corrected by some other process, such as editing, into our more constrained set of variations.

\section{Our approach}
We seek to differentiate between the ability to select corrections from a constrained set within a stable context, and the ability to detect errors when nearly every word is a potential error.  To do this we first construct a corpus of errors out of naturally occurring text by randomly substituting words with other words that are close in spelling.  We demonstrate that if the problem is constrained to discriminating between the original word and a single random variation of it within the context of a sentence that is otherwise assumed to be correct, a standard trigram model performs at high accuracy.

We then construct a hidden Markov model that uses a trigram model for transition probabilities and a noisy channel model for emission probabilities.
We use the Viterbi algorithm to find the most probable sequence of word variations for each word in each sentence of our corpus.
This structure allows every word to be a candidate for correction to any of its allowed variations even when the variations interact.
The results shown below demonstrate that many words originally in the text are not the most probable of their variations, when probability is given by a trigram model.
%Moreover, no simple approach to detection thresholding can accurately separate correct words from errors using trigram probabilities, although such an approach can significantly improve speed.

%XXX Given this framing, it now seems necessary to test directly how well trigrams would perform if detection were given, i.e. we knew what words were errors, and just chose the most probable variation in its place.

%\chapter{Methods}

\section{Evaluation}

\subsection{The problem of evaluating unsupervised learning}
We have cast the problem of real-word spelling error detection and correction as a task involving two independent models.
We learn a model of natural language in an unsupervised fashion, in this case a word trigram model.
We also design an algorithm that uses that model
to recognise a specific kind of ill-formed text by finding similar text that fits the model better.
Here we use a noisy channel model in this role.
Evaluation of the whole task necessarily evaluates both parts.

Unsupervised models are challenging to evaluate.
\citet{smith2012adversarial} analyses evaluation practices in NLP, drawing distinctions between intrinsic, extrinsic, and perplexity evaluations, and introducing a perplexity-like evaluation framework based on adversarial roles.
We summarise this analysis here.
Intrinsic evaluation measures how well a model can replicate previous analyses, usually human annotations.
This is ill-suited to unsupervised learning, because it restricts learning to a pre-defined structure, and unsupervised models often have learning the structure itself as part of their goal.
Extrinsic evaluation measures performance on a downstream task.
Unfortunately, it can conflate the evaluation of a model with the evaluation of the architecture in which it is embedded.
Perplexity measures how well a model predicts subsequent text,
but it is constrained to models that assign probability and it is highly sensitive to differences that may be unimportant in practice.
For example, small differences in smoothing approaches may result in large differences in perplexity that would not be reflected in performance.

In the proposed framework, instead of evaluating a language model directly on the probability it assigns to test data, as in perplexity, a minimal task is proposed.
That task is to discriminate between an instance of natural text, and the same instance that has been subtly altered.
We will refer to this as the \emph{binary original text discrimination} (BOTD) task.
Because this measure is dependent on the quality of the alterations, which is in turn dependent on current models, the framework makes explicit an interdependence of measurement between language models and problem spaces.
Our evaluation for spelling correction is interpreted with respect to these insights.

\subsection{Adversarial role evaluation}
Intuitively, we might like to evaluate spelling error detection and correction extrinsically, by measuring how well it can match corrections to textual data containing naturally occurring misspellings.
This would require as a resource a data set that includes spelling mistakes and their corrections.
Such resources are difficult to find.
\citet{zesch2012measuring} uses Wikipedia revision history as a source of such corrections.
Our approach instead uses artificially created errors.
An existing corpus that is assumed to contain few or no errors is transformed by replacing some words with variations of those words.

Criticisms of this approach include lack of realism, specifically, inclusion of errors that are unrealistically easy to detect and correct (\citet{zesch2012measuring}), and missing or misweighting error types that actually occur.
Artificial errors may also include those that are impossible to correct, because they are indistinguishable from the original in their context.

However, these potential problems are a property of the error generation algorithm, and the state-of-the-art of language models.
Errors that are easy to detect and correct are easy only insofar as we have models that can account for them.
By analogy with the adversarial framework described above,
we conceive of evaluation of spelling error detection and correction as consisting of three interdependent roles:
the transformation algorithm that creates the spelling errors,
the language model that scores instances on their well-formedness,
and the spelling correction algorithm that uses the language model to propose corrections.

Analysis of the mistakes of the model plays a central role.
Transformation algorithms can be improved by iteratively focusing on including error types that are not easily corrected by current language models.
This in turn would stimulate the development of better language models.
In other words, choosing the kind of transformations to apply to the corpus is itself a research question.
We argue that using artificial errors is a strength when it exposes errors that are easy to detect and correct.

There are other strengths of using artificial errors.
Foremost, artificial errors are of unlimited supply.
Also, they can be created in any genre of text.

We make use of the BOTD task as a way to pinpoint limitations of the language model.
Asking our trigram model to distinguish between the original text and a transformation of that text without supplying the added information of which text was observed, defines a ceiling for the correction algorithm when it uses that trigram model.
Failures to correct sentences with those transformations that cannot be discriminated from the original by the trigram model in isolation reflect the limits of the selection ability of trigram model, and not the correction technique.

Because this task singles out selection,
it also effectively brings into focus the detection capability of the corrector.
The decision of which word's correctness is in question has already been made in the BOTD task, so no false positives can be made.
Also, a word error that can be successfully recognized in the BOTD task, but that is not correctly corrected by the corrector, was either not detected or miscorrected into yet another word.
If the BOTD task accuracy is high, then most of the cases will be of the former type.

\subsection{Generating the error corpus}
We contrast the method used to generate the test corpus here with two similar methods that have been used previously.

The test set of \citet{mays1991context} was derived from a set of 100 newswire sentences, none of which contained words outside of their 20,000 word vocabulary.
It contained those sentences, along with every possible one-word variation of those sentences.
A variation was constrained to replacing one word in the sentence by another word in the vocabulary of DL-distance $1$ from the original.
The authors evaluate this test set using the same scoring system as the BOTD task: The sentence as a whole is scored as correct if and only if it was restored to the original.
Unlike the BOTD task, there is now a much larger set of alternative sentences to choose among.
Also, one sentence, the observed one, is privileged in that it is given a boost (or decrement) in probability with respect to the others,
to reflect the probability of noise from the channel.
Because the choice now includes the decision of which word or words to vary, detection gains some prominence.
For this reason, accuracy is no longer an appropriate measure: it conflates false negative detections with miscorrections of the right word.

The ability to evaluate detection in this setting is still limited.
Because the set contains a relatively high rate of errors,
the probability of there being an error in any given sentence we attempt to correct is very high.
Even though there are more words from which a variation is being considered, the assumption that one of these should be changed will still usually be correct.
So, an algorithm that corrects frequently may appear to detect errors with high precision, but this may translate to low precision on data with more moderate error rates.
To clearly evaluate detection, the error rate must be lower.
Relatedly, the constraint of at most one erroneous word per sentence allowed an algorithmic solution that took advantage of that constraint, and could not be applied when considering sentences with multiple errors.

The method for generating test sets reported by \citet{hirst1998lexical}, \citet{hirst2005correcting}, and \citet{wilcox2008real} addresses the inflated error rate by creating a single copy of each sentence in the source.
Only some words in some of these sentences had errors applied to them.
In a corpus of 500 Wall Street Journal articles,
error transformations were applied to one in every 200 words.
In the former two studies, only non-stop words in the vocabulary were eligible for transformation.
While this method provided a more desirable error rate, it continued to constrain the number of errors per sentence to at most one in almost all cases.
Certainly no sentences could have adjacent errors.

There are two differences between the method used to generate the error set here and the latter method.
First, the source text came from Wikipedia articles.
This genre of text has a more varied vocabulary, and is less consistent in style and spelling than the WSJ.
This warranted some special treatment in sentence segmentation as detailed in the documentation of the code.
It also had implications for word type, as discussed in section 8: there are spelling variations and misspellings in the corpus itself.

The second difference is that errors were inserted probabilistically.
With a 1 in 200 \emph{chance}, every word occurring in the text was considered for transformation.
It was transformed if and only if it was in the vocabulary and had at least one variation of DL-distance $1$ also in the vocabulary.
As before, among those variations, one was selected uniformly randomly.

\subsection{Evaluation Measures}

\subsubsection{Precision and recall defined for detection and correction}
Normally, to calculate precision and recall, we count true positives, false positives, and false negatives.
This terminology is straightforward for detection.
For correction it is somewhat confusing.
To illustrate, consider the possible classes of correction represented by the following tuples of (original, error, correction):
\begin{quote}

$(x,x,x)$ True Negative ($TN$)

$(x,x,y)$ False Positive ($FP$)

$(x,y,x)$ True Positive ($TP$)

$(x,y,y)$ False Negative ($FN$)

$(x,y,z)$ Detection True Positive, Miscorrection ($MC$)
%Correction False Positive \emph{and} False Negative
\end{quote}

Correction precision measures the proportion of proposed corrections that were correctly corrected.
Since $(x,y,z)$ has been corrected, but not correctly, it is a false positive.
$P = TP/(TP+FP)$.
We interpret recall to measure the proportion of all errors that were correctly corrected.
Since $(x,y,z)$ is an error, and it is not correctly corrected, it must be counted as part of the space of errors along with False Negatives,
even though $(x,y,z)$ cannot itself be considered a negative (a correction was indeed proposed).
$R = TP/(TP+FN+MC)$.

Correction accuracy is given by $(TN+TP)/(TN+FP+TP+FN+MC)$.

\section{The correction process}

We specify the following components:
a language model, a vocabulary, a confusion set generation algorithm, and a correction algorithm that selects the best fitting variation.

\subsection{Vocabulary, trigram model, and confusion set generation}
For the vocabulary, we first tokenised the training set by separating at all space boundaries.
Within space boundaries, we also isolated all punctuation to single characters with the following exceptions:
We followed the convention of splitting contractions into two parts, the second of which keeps the apostrophe.
Inter-numeric commas and periods stayed token-internal.
Ellipses composed of periods were kept together as a single token.
Periods were kept on abbreviations and initials, insofar as they were recognised by our customised version of NLTK's sentence segmenter \citep{bird2009natural}\footnote{
NLTK's \emph{punkt} segmenter learns to recognise abbreviations, to overcome the problem of mistakenly assuming a sentence has ended at an abbreviation period, and proper nouns, to help identify capitalisation patterns that do not mark the beginning of sentences.
Our customisation is a combination of setting existing NLTK parameters to cope with the inherent inconsistencies in Wikipedia's word forms, and an added heuristic to address a specific, common error we had noticed involving mistaken splitting of sentences between intials in names.}.
Digit strings were replaced with a regularising token based on the number of digits in the string \footnote{
We reason that some lengths of digit string tend to occur as semantically substitutable types, such as dates, and the cent portion of a monetary amount.  However, we did not test this against other options.
}.
We then selected all tokens occurring more than once to form a base vocabulary.

The trigram model was made using this base vocabulary and the SRILM toolkit \citet{stolcke02srilm}.
For smoothing, we used SRILM's implementation of the modification of Kneser-Ney discounting \citep{kneser1995improved} described by \citet{chen1999empirical}, and the backoff method of \citet{katz1987estimation}.
We also used an unknown-word token to estimate the probability of previously unseen words, by assigning the probability mass of any word occurring only once in the training data to this token (also implemented by SRILM).

A \emph{real-word} vocabulary was then derived from the base vocabulary by excluding all tokens that contained no letters, or did contain symbols other than apostrophes or periods.

%\subsection{Confusion set generation}
We generated confusion sets by finding all tokens of DL-distance 1 from the original that occurred in the real-word vocabulary.

%There are two approaches to finding all vocabulary words within a DL-distance $d$.
%One is to generate all strings of distance $d$ and then check to see which of those strings are in the vocabulary.
%The other is to calculate the DL-distance of every pair of words in the vocabulary.
%
%For $d=1$, the former is preferable.
%Suppose your character set is of size $C$.
%A token of length $n$ will have $n$ possible deletions, $n-1$ transpositions, $n{\times}(C-1)$ substitutions, and $(n+1){\times}C$ insertions.
%$C \gg n$.
%Generating and testing these possibilities requires $O(Cn)$ string transformations and table lookups.
%On the other hand, comparing a pair of tokens of length $n$ and $m$ is an $O(nm)$ operation, and it must be done $O(C^2)$ times to get all the pairs.
%
%However, for a general $d$, to use the generate-and-check strategy we must take all of the strings from the search in $d-1$ (whether or not they were in the vocabulary) and compute a further $O(Cn)$ search for each of those.  This grows exponentially, and is already as complex as calculating the distance for each pair at $d=2$.
%At $d>1$, therefore, it is preferable to compute the distances in advance, and store them.
%
%Because we use only DL-distance $1$, it is efficient enough to compute them during processing, though it would probably improve speed to store them.

\subsection{Distribution of errors}
In this set of experiments we consider all word variations of DL-distance $1$ to be equally likely.
Although there exist models (\citet{kernighan1990spelling}) intended to capture the likelihood of specific letter-to-letter errors, we do not incorporate them.
Such models assume homogeneous input methods.
Moreover, the methods used to create them have limitations, and there is no convenient way to test their applicability.
The same problem of lack of intrinsic or extrinsic evaluation resources we find in the spelling correction task itself, also applies to the task of generating letter error models.
From a pragmatic standpoint,
for the correction model to be improved by letter error rate distributions,
the error set creation would also have to reflect those distributions.
However, adding more components to the current model would only obscure the analysis of the multiple aspects of it already under consideration.

\subsection{The correction algorithm}
In the previous models that used this approach it was usually assumed that at most one error could occur in a single sentence.
This enabled the strategy of simply comparing the probability of the original sentence with that of every sentence that could be formed by replacing one word with its spelling variations.
\citet{wilcox2008real} also reported some attempts to correct multiple errors by breaking the sentences into smaller \emph{windows}, and either correcting those windows in isolation (tiling), or sliding them across the sentence.
This method still could not model errors within one window length of one another.
The current corpus generation allows multiple errors to occur in one sentence, regardless of sentence length.
Although this happened only 101 times out of 38,710 sentences,
the possibility motivated a dynamic programming approach to correction.

The models we used for correction are varieties of the hidden Markov model (\emph{HMM}).
The hidden states correspond to the intended words that resulted in the observed words.
Transition probabilities are given by a smoothed trigram model trained on the training set.
Emission probabilities cover the set of variations of the word represented by the state.
The probabilities themselves incorporate a parameter ${\beta}$ intended to reflect the noisy channel parameter ${\alpha}$ (the probability of no error) and are not smoothed;
there are no possibilities admitted except the variations of the observed word, as given by the variation generating algorithm (DL-1).
Then we use a Viterbi search to find the most probable sequence of intended words.

%XXX diagram?

Comparing this with the algorithm from \citet{mays1991context},
we now have in our set of alternative sentences every combination of every variation of each word in the sentence.
The ${\beta}$ parameter serves to bias not only the observed sentence, but every sentence to the degree that it has fewer variations from the observed sentence.
However, strongly interacting variations may overcome these biases.

\section{Experiments and results}

\subsection{The binary original text discrimination task}
As described above, we use the BOTD task to provide a ceiling for our results, and to isolate the performance of the spelling corrector from the selection ability of the trigram model.
On this task, for the error corpus with an error rate of 1 in 200, the trigram model correctly chose the original sentence with an accuracy of 0.949.
This demonstrates that a trigram model performs selection well --- only about 5\% of randomly chosen variations were scored higher than the original.
Based on this result, we expect that given correct \emph{detection}, few errors will be miscorrected.

We also generated a corpus with an error rate of 1 in 20, which we will use to demonstrate that the relationship between propensity to correct and precision is dependent on error rate.
For this corpus, the accuracy rose to 0.964.

This difference probably comes from a higher occurrence of sentences with multiple errors;
more errors per sentence gives more chances to reject the sentence as a whole.
In contrast to the 101 sentences out of 38,710 containing multiple errors that occurred with the 1 in 200 error rate, using a rate of 1 in 20 resulted in 5,084 sentences with multiple errors.

\subsection{Trigram Viterbi}
In this experiment we modify a standard HMM for trigram probabilities, such that
the states for position $i$ represent not only the intended word at position $i$, but also the intended previous word.
The selection of a word variation at position $i$ is determined by its probability in the three trigrams it participates in:
$P(w_i|w_{i-2},w_{i-1})$, $P(w_{i+1}|w_{i-1},w_i)$, and $P(w_{i+2}|w_i,w_{i+1})$.

Let $Var(w_i)$ be the set of possible intended words at position $i$.
Then at position $i-1$, there will be $|Var(w_{i-2})|{\times}|Var(w_{i-1})|$ states, and
at position $i$, there will be $|Var(w_{i-1})|{\times}|Var(w_i)|$ states.
Transitions from position $i-1$ to position $i$ only occur when the second word of the first matches the first word of the second.
This yields $|Var(w_{i-2})|{\times}|Var(w_{i-1})|{\times}|Var(w_i)|$ transitions between the states at $i-1$ and the states at $i$.

For each state at position $i$, the Viterbi algorithm commits to the best incoming state from $i-1$.
Because the search space is so large, we then prune the states at position $i$ to the $t$ most probable, where $t$ is a tunable parameter.

This algorithm was repeated for three values each of $t$ and ${\beta}$, for the error rate of 1 in 200 (Table 1).
For the error rate of 1 in 20, we computed the results only for $t=3$ (Table 2).

%\subsection{Results}

\begin{table}[h]
  \begin{tabular}{llccccccc}
    & & \multicolumn{3}{c}{detection} & \multicolumn{3}{c}{correction} & correction accuracy\\ \cline{4-4} \cline{7-7} %\cline{9-9}
    $t$ & $\beta$ & P & R & F & P & R & F \\
    \hline
    \multicolumn{9}{l}{3} \\
    & 0.95 & 0.176 & 0.771 & 0.287 & 0.165 & 0.722 & 0.269 & 0.936\\
    & 0.995 & 0.375 & 0.587 & 0.457 & 0.361 & 0.565 & 0.440 & 0.962\\
    & 0.9995 & 0.611 & 0.412 & 0.492 & 0.600 & 0.404 & 0.483 & 0.981 \\
    \multicolumn{9}{l}{9} \\
    & 0.95 & 0.171 & 0.777 & 0.281 & 0.161 & 0.729 & 0.263 & 0.939 \\
    & 0.995 & 0.367 & 0.593 & 0.454 & 0.355 & 0.574 & 0.439 & 0.967 \\
    & 0.9995 & 0.603 & 0.415 & 0.491 & 0.593 & 0.408 & 0.483 & 0.984 \\
    \multicolumn{9}{l}{27} \\
    & 0.95 & 0.170 & 0.777 & 0.279 & 0.160 & 0.730 & 0.262 & 0.939 \\
    & 0.995 & 0.365 & 0.593 & 0.452 & 0.353 & 0.573 & 0.437 & 0.967 \\
    & 0.9995 & 0.603 & 0.415 & 0.491 & 0.592 & 0.408 & 0.483 & 0.984 \\
  \end{tabular}
  \caption{Error rate 1 in 200 ($\alpha=0.995$). $t$ is the number of paths kept per position, $\beta$ the noise probability parameter, and P,R,F are precision, recall, and F-measure.}
\end{table}

\begin{table}[ht]
  \begin{tabular}{cccccccc}
    & \multicolumn{3}{c}{detection} & \multicolumn{3}{c}{correction} & correction accuracy \\ \cline{3-3} \cline{6-6} %\cline{8-8}
    $\beta$ & P & R & F & P & R & F \\
    \hline
    0.95 & 0.690 & 0.771 & 0.728 & 0.649 & 0.726 & 0.685 & 0.941 \\
    0.995 & 0.868 & 0.614 & 0.719 & 0.836 & 0.591 & 0.692 & 0.936 \\
    0.9995 & 0.945 & 0.423 & 0.588 & 0.924 & 0.418 & 0.576 & 0.978 \\
  \end{tabular}
  \caption{Error rate 1 in 20 ($\alpha=.95$); $t=3$ is the number of paths kept, $\beta$ the noise probability parameter, and P,R,F are precision, recall, and F-measure.}
\end{table}

From the tables we can see that given detection, correction accuracy is very high, as expected.
Detection itself has much poorer results.
To get only slightly more than three-quarters of the right words even under consideration,
we also mistakenly change more than four innocent bystanding words for every real error we detect.
%xxx not quite right: this kind of assertion requires false negs? more than four out of five innocent bystanding words.
On the other hand, giving the observed words enough benefit of the doubt to reduce the false positives to only a little less than two for every three correct,
reduces our recall so much that we are detecting only half of the errors.

As we anticipated, allowing a higher error rate increased precision immensely at a given recall rate, because the opportunity for false positives is so much less.

\section{Discussion}

\subsection{Effects of the genre}
The property of Wikipedia articles that makes it the most different from newswire is its vocabulary.
We measured this using the first 14 sections of the WSJ corpus from 1987, which collectively has a similar number of tokens (as estimated by \emph{wc}) to our Wikipedia training set.
We segmented and tokenised this data in the same way as we did our training data, and then counted the type frequencies.
While the WSJ articles contained only 44,035 distinct types, about 65 percent of those were hapax legomena.
The Wikipedia training set had 88,078 types, 51 percent of which were hapax legomena.
Adding more WSJ articles to double the tokens increased the type count only to 58,477, with a small increase in percent hapax legomena to 68 percent.

The difference in type counts may be partly because Wikipedia is encyclopedic, and therefore covers a much wider variety of topics than the Wall Street Journal.
It is also probably due to the much greater number of authors, and the concomitant lack of standards for spelling.
We attribute the smaller percentage of hapax legomena in the Wikipedia data to the fact that new ideas or even names would tend to be explained (and therefore repeated) rather than mentioned.
We also expect Wikipedia to contain a higher rate of non-word spelling errors.

Insofar as it is correct that the Wikipedia text contains legitimate spelling variations of the same types, and also spelling errors, this violates the assumptions of our algorithm\footnote{
This effect may be more common in Wikipedia than in newswire, but newswire is not immune to it. For example, the 1987 portion of the WSJ corpus contains two occurrences of the token `billlion', and six of `milllion'.
}.
This could cause failures of the correction algorithm.

For example, suppose there were a spelling error that occurs just a few times in the corpus.
This erroneous token will be used when generating the error corpus.
It will usually be easy to correct, because the trigram model will estimate it to have low probability.
However, if it happens to occur in a similar context to the one in which it occurred in training, and if the target correction word did not occur in that same context sufficiently often, the misspelling will actually be assigned a higher probability than the correct spelling.
For example, our training text contained two occurrences of the misspelling `wast' for `was'.
Two occurrences was our threshold, so `wast' made it into the vocabulary.
One of the two contexts was `... subject matter wast influenced by...'.
The bigram `matter was' also occurred only once.
This means that any text containing `matter wast influenced' will score a higher probability than `matter was influenced' regardless of the token before or after these three.

If a misspelling is quite common, or if it is actually a spelling variation (the distinction may be philosophical), this effect would be more pronounced.
The variation that is chosen will reflect the idiosyncrasies of the contexts it appeared in in training.
For instance, `travelling' and `traveling' occur with about the same frequency in our training set; 45 and 43 times each.
While they share some surrounding words, they also have differences.
For example, `travelling' appeared as an adjective before these nouns: `Australians', `bachelor', `communities', `keyboard', `life', `opera', `post', `support', and `way',
whereas `traveling' appeared before these nouns: `champion', `conditions', `exhibition', `odyssey', `show', and `troop'.
The prepositions following `travelling' were `at', `between', `from', `in', `on', `past', through', `to' and `via',
whereas `traveling' was followed by the prepositions `across', `back', `between', `by', `for', `in', `into', `off', `on', `though', `to', `until', `with', and `within'.
These differences may indicate true correlations in style, or they may simply be artefacts.

In order to avoid detecting an error when one of these occurs,
we would like to recognise them as being of the same type, perhaps through some process akin to those used to detect near-synonymy.
This would allow an enforcement of consistency at the level of the article or corpus.

\subsection{Effects of the models}
As shown above, only about five percent of the transformations in our error corpus were more likely in their contexts than the original as estimated by the trigram model.
This means that most of the detection errors are attributable to the corrector.
We discuss these first.

Within the corrector, the variation generator has been designed to guarantee that the correct variation is included for consideration.
So the task of generating variations has been removed from evaluation.
This means recall can potentially be very high.
The fall in precision at high recall reflects the rate of words that occur in the text that are not as probable as alternatives.
Even at a low rate, this can have a large effect on precision.

Low precision may pose a usability problem, as a high rate of false positives would be expected to undermine confidence in a spelling corrector and to be frustratingly distracting.
In word prediction, on the other hand, more options are less likely to interfere.

\subsection{Results}
The results were lower than expected, given the stronger results reported in \citet{wilcox2008real}.
We attribute the difference to the genre.
Because Wikipedia articles are explanatory, and the topics sometimes obscure, terms in an article tend to be both rare and repeated.
Thus our algorithm tends to wrongly correct rare words, such as names, repeatedly.

%For example, table 4.1 shows the most common false positives.
%It includes, for example, `Tawe' being corrected to `Tame'.
At lower values of ${\beta}$, false correction of names is more acute.
For example, with ${\beta}=0.995$ (not shown), the most common false positives
included `Riddle' being corrected to `Middle', `Beatty' to `Beauty', and `Lucia' to `Lucian'.
Names are likely to occur many times if they occur once, and be mistakenly corrected every time if at all.
We could teach our system to recognise named entities and not correct them,
but a more appropriate approach might be to adapt our probability estimates to reflect recent words.
% (see Chapter 5).

This latter solution might also mitigate the other common kind of false positive:
words that are uncommon enough that a random increased frequency in the training data of one variation over another strongly biases the estimation of their probability in unseen contexts.
An example of this is `engines' being corrected to `engine' many times in an article about jet engines, even though \emph{within} the article, `engines' occurred 79 times.
This is related to the valid variations of type in the `travelling'/`traveling' example, and highlights a difficulty in distinguishing types:
We expect there to be a trade-off in discovering type variations and detecting spelling errors.
Indeed, another kind of false positive in our list was the rejection of spelling variations of `Istanbul' and `Ictimai'.
This problem could be solved with world knowledge, but word type recognition would be a more elegant solution.

Finally, the correction of `billion' to `million' exemplifies a kind of error that would be difficult to avoid using this method without incorporating world knowledge, and perhaps impossible even then.

False negatives, on the other hand, tended to be common words.  The word `and' was left unrestored particularly often in this experiment.
This may be because `and' has a diffuse distribution of words it can occur next to, making it less likely than average to occur next to any specific one.

The most common correctly corrected word was `the' by far, though this may be mostly a reflection of its being most commonly transformed, and its very high probability.
We did not compare words for these effects.

Another unexpected result was that more pruning of the HMM states did not result in unambiguously worse results.
Instead, more pruning improved precision slightly more than it degraded recall, for a net slight \emph{improvement} in F-measure.
Again, this reflects the problem of too many false positives that occurs when the number of choices is high.
In the section 2, we described how systems that over-correct would be expected to perform better on a corpus with a higher error rate.
We demonstrate this by including results on a corpus with a higher error rate (Table 2). 

%\section{Surprise results}
%The threshold method of detection did improve performance very significantly, while affecting the results only negligibly.
%Contrary to our hypothesis, using the back-off weight gave no advantage in results over the simple threshold by itself.
%Moreover, because of the extra look-ups required to find the back-off weight, performance improved more slowly.

\section{Future work}
In contrast to speech recognition systems, most NLP systems that operate on written text take observed words as given.
For example, a typical parsing system would not propose a parse that considered the possibility that a word of its input were a misprint, or a variation of a different word, even if there were a word close in spelling with a much more plausible parse.
If such a system addressed spelling variations at all, it would typically be done separately from and in advance of parsing, in a pipelined fashion.

However, as our source texts have become increasingly diverse, and standardised spelling is increasingly a genre-specific feature, NLP systems have more to gain from a flexible definition of word \emph{type}.
That is, rather than considering every orthographically distinct word form as a separate vocabulary item, it should be helpful to recognise that some forms are variations of the same type.
This has motivated much work in \emph{text normalisation}, the task of assigning all spelling variations of a word to one standard representation.
If we assume that the observable forms of different types may overlap,
then text normalisation can be considered to subsume spelling correction,
because the type of an observed word is then ambiguous.

We therefore argue that modern processing of written text would benefit from integrated \emph{word (type) recognition} in analogy with speech recognition systems.
That is, for text processing in which word forms may vary (either as a result of error, or as a feature of the language), we expect that language modelling, parsing, and other tasks that either measure text plausibility or are steps in natural language understanding will be more efficient and accurate if the word type recognition is computed during those tasks rather than before.
The system presented here performed spelling correction by integrating word form recognition into a trigram language model using a hidden Markov model.
The same method could be used to perform trigram modelling with implicit text normalisation.

We also propose that for both spelling correction proper, and the text normalisation problem,
an adaptive approach that tracks corrections may help avoid pitfalls.
For example, many attestations of a name or a singular vs. plural distinction that repeatedly goes against the expectations of a model might be a good cue to change the model, at least within a given article or defined context.
Tracking corrections could also help enforce consistency of type, such as the `travelling'/`traveling' example.

However,
for any such system to do more good than harm,
the problem of over-detection must first be resolved.
The problem of selecting among multiple simultaneous variations turns out to be much harder than selecting between an original word and a single variation of it.
The difficulty is a problem in detection---the choice of which words to keep, and which to correct---as demonstrated by the high rate of correction given detection.
Although n-gram models are powerful statistical tools, the structure of language that allows errors to be detected easily by speakers is not well captured by them.

%a stronger probability model than we used here will need to be incorporated.

%\chapter{Conclusion}
%On the other hand, in some cases, for example if the text was given my a foreign speaker, a systematic error may be safely corrected every time.

%Most words that actually occur in written text are not the most probable word.
%If that were the case, no information would be transmitted.

%In this paper we show that by allowing a more realistically open set of possible errors than in some previous research, trigram models cannot simultaneously reliably find true errors without introducing many more, at least not when used in the obvious sequential way without added structure.
%This is true even if probable words are excluded from consideration in advance.

If, according to a trigram probability model, 1 in 20 naturally occurring words are not the most likely among close variations of them,
then a typical correct sentence will have at least one attractive miscorrection.
If real errors occur only, for example, 1 in 200 words,
then we might expect nine false positives for every true positive, when looking for errors aggressively.
Therefore, while $95\%$ accuracy sounds high when evaluating a language model, it is not sufficiently high to perform well when detection of sparsely occurring errors is the task.

Because detection is the critical performance issue we encountered, we expect that future work will have to address it.
Natural language comes with a high level of built-in redundancy.
A model that can more effectively tap into this inherent structural property of language should be much more adept at detecting real word errors.

\bibliographystyle{plain}
\bibliography{applying}

\end{document}